\documentclass[conference]{IEEEtran}
\IEEEoverridecommandlockouts
\usepackage{cite}
\usepackage{amsmath,amssymb,amsfonts}
\usepackage{algorithmic}
\usepackage{graphicx}
\usepackage{textcomp}
\usepackage{xcolor}
\usepackage{url}
\usepackage{hyperref}
\usepackage{color,soul}
\graphicspath{ {./figures/} }

\def\BibTeX{{\rm B\kern-.05em{\sc i\kern-.025em b}\kern-.08em
    T\kern-.1667em\lower.7ex\hbox{E}\kern-.125emX}}
\begin{document}

\title{Evaluation of computer networking methods for  interaction with remote robotic systems
\thanks{The research leading to these results has received funding from the European Union’s Horizon 2020 Research and Innovation Programme under grant agreement No 731103, "EUMarineRobots" and Erasmus+ Strategic Partnership "IMPACT - Intelligent Marine systems
a Pathway towards sustAinable eduCation, knowledge and
empowermenT" 2018-1-DE01-KA203-004259}
}

\author{\IEEEauthorblockN{Peter-Newman Messan, Szymon Krupi\'{n}ski, Francesco Maurelli}
\IEEEauthorblockA{\textit{Marine Systems and Robotics} \\
\textit{Jacobs University Bremen}\\
Bremen, Germany \\
\{p.messan; s.krupinski; f.maurelli\}@jacobs-university.de}
\and
\IEEEauthorblockN{Guillem Vallicrosa, Pere Ridao}
\IEEEauthorblockA{\textit{Computer Vision and Robotics Institute (VICOROB)} \\
\textit{Universitat de Girona}\\
Girona, Catalonia, Spain \\
name.surname@udg.edu}
}

\maketitle

\begin{abstract}
Use of robotic infrastructures can significantly increase with remote access. This would open up the possibility to use costly equipment without the need to buy them, or to simply access those assets remotely when actual travel is not possible or recommended - for example in pandemic times. In this paper we present an analysis of several networking techniques which  allow remote robotics operations, alongside with experimental results with distance ranging from hundreds of meters up to thousands of kilometers. 
\end{abstract}


\section{Introduction}
The use of robotics is decisively increasing worldwide. In 2018 the International Federation of Robotics analysed the growth of the robotics density, defined as number of robot units per 10,000 employees~\cite{ifr2018}.
Depending on the world region considered, the growth registered varied from 5\% in Europe up to 9\% in Asia, through 7\% in the Americas. Africa is however not included in this statistic. Despite the growth in the use of robotics in Africa, especially for drones and education, as outlined by Vernon~\cite{Vernon2019}, there is still a substantial gap compared with other more industrialised areas of the world. In some cases the fear of a negative impact of robotics and automation in the job market slow down the overall process, though studies have shown that new technologies do not represent a threat~\cite{Gaus2019}.
One of the key barriers for more widespread use of robotics, both in industrial settings and in research, is accessibility. Robotics systems can be very expensive, and not all universities or companies have the means for such investment. At different scales, from accessing a single simple robot up to accessing a complex complete robotics infrastructure, this is a challenge for most researchers worldwide, and not just in one geographical area of the world. Considering that, especially in research, robotics infrastructures at each institutions are rarely used full time, as they are often required only for the experimental section of the work, sharing robotics infrastructures is a possible alternative path, which would allow more users to use the available facilities. In this way, the accessibility barrier is significantly lowered, no high up-front investment cost is needed for the user and at the same time the provider would make use of the robotic systems for a more extended time. Supporting the idea of sharing and collaborating, the European Union has launched several infrastructure initiatives, to open up key national and regional robotics research infrastructures (RIs) to all researchers, from both academia and industry, ensuring their optimal use and joint development to establish a world-class robotics integrated infrastructure. One of these examples is the EUMarineRobot project\footnote{https://www.eumarinerobots.eu/}, focused on Marine Robotics.
This approach still presents however some barriers: the travel to the robotic infrastructures. Whilst this is sometimes a negligible expense compared to the asset cost, depending on the location of the infrastructure, this can still represent a significant barrier, not only from the financial perspective, but also considering physical accessibility (e.g. visa requests). Whilst remote control of robotic systems was already progressing until last year, the Covid19 pandemic has created a unique push for remote working, remote collaborations and therefore also for remote access to robotic infrastructures. Additionally, the need to access robotic systems from a remote location is not limited to researchers and workers in the specific field, but it became a necessity also to support educational activities, and allow students to interact with those systems in a setting as similar as possible as the pre-Covid laboratories~\cite{Birk2021_edu,Maurelli2021_rie}.
With an increase of network-based work and communications, aspects of security need to be considered. Whilst for some simple use cases or demonstration of remote control of a robot a simple \texttt{ssh} to the public IP of the robot may be sufficient, in many cases, this is not so. Many robots are simply connected to wifi routers, thus hidden from other devices on the internet not in the same Local Area Network. This implies connecting to the public IP means connecting to the router assigned that IP, and not the robot. This situation is even more difficult when dealing with large, heavily insulated network architectures such as university and enterprise networks, like the one we have at Jacobs University Bremen. Such setups may have a fixed range of public IP addresses assigned to one or more routers with a protective firewall, and all other routers inside the establishment operate under that range of IP addresses, and so do their connected devices. Thus, trying to access a specific device inside such a network using only the public IP is next to impossible, and requires a workaround that needs to be both secure and highly available for research and general purposes.
The objective of this paper is therefore to present a comparison among various network approaches in order to facilitate the choice of the right infrastructure for researchers interested in remote collaboration and remote access to robotic infrastructures.
This paper is organised as follows: Section~\ref{s:rw} presents the related work; Section~\ref{s:na} outlines the various network approaches analysed; Section~\ref{s:er} presents the experimental framework and the results of the tests in the different network approaches. Finally, Section~\ref{s:cfw} summarises the main aspects of this work, outlining future work to complete the analysis.

\section{Related Work}
\label{s:rw}
In 1999, Fitzpatrick presents what is considered the first public robotic remote control: in April 1997, at the ICRA conference, an operator controlled the motions of a robot more than 1,500 kilometers away~\cite{Fitz1999}. According to Fitzpatrick, the teleoperation of the robot was a major step for the burgeoning field of robotics, and it was considered  a giant leap for telecommunications. It added a previously unexplored dimension to the Internet-remote control and provided new horizons for other applications in long-distance learning, research opportunities, and hands-on experimentation. In the following years, several work appeared on robotics remote control via the internet. Saucy and Mondada report about  remote accesses of a mobile robot by the general public over a period of one year~\cite{saucy2000}.
The system was based on a Microsoft Personal Web Service using Windows95, using CGI scripts to perform the tasks. The robot was connected via serial link RS232 and a shared memory was used in order to share images from the robot camera. This was a revolutionary example of how anyone on the internet with no specialised knowledge was able to interact with a robotic system, overcoming all the limitations and challenges posed by the time, and requiring frequent system re-start.
Luo and Tse~\cite{Luo2000} propose a solution for the remote behavior-programming control of intelligent mobile robots through high-latency Internet. The components
for providing autonomous capabilities of a mobile robot are grouped into motion planner, motion executor, motion assistant, and rule-based behavior arbitrator. As internet speed and reliability was very different than in current days, the work focused on providing task-level commands, sent to the robot via radio Ethernet, rather than direct remote control. The emphasis on time delay over internet has been the emphasis of many of the early work in the field, as also addressed by Han \textit{et al.}~\cite{Han2001}. They proposed a novel internet control architecture for an \textit{internet based personal robot} (IPR). Uncertainty and time delays were the main focus of the work, with an architecture robust to internet time delay. This is achieved thanks to \textit{command filter}, which can recover the information loss of control commands caused by internet time delay. A few years later, Wang \textit{et al.} propose a new server-decentralised internet architecture, based on Jabber for robot remote control~\cite{Wang2006}. Four components of the architecture are presented: operators,
robots, transfer servers and data-keepers. The robot-controlling data and robot state data are
expressed in XML and delivered to the addressed
robot/operator through XML streams. With the popularisation of smartphones, efforts went into remote-control via mobile applications. Nádvorník and Smutný present the design and realisation of
a mobile application for the Android operating system~\cite{Nad2014}. This is focused on manual control of mobile robot using wireless
Bluetooth technology. The application allows the robot control interaction with the display, or voice. In order to work for long-distance remote control, this and similar works would need to be extended with an additional internet-based layer.
The increased internet speed and reliability of the connections have been key enabler to applications of telerobotics in medicine. Su \textit{et al.} present \cite{He2018} the remote control of a minimally invasive laparoscopic surgery robot. A telesurgery
communication protocol was set up based on Transmission Control Protocol/Internet Protocol. The stereo images of
laparoscopy were transmitted by a hardware-based H.264 encoder/decoder. A demonstration environment of robot assisted remote minimally invasive surgery was successfully demonstrated between two locations located at about 150km distance. Progress in communication with 5G technology also pushed research in remote robotics forward. Tsokalo \textit{et al.} presented the key requirements of a remote robot control 5G use case, considering the
accuracy and real-time reaction time requirements~\cite{Tsokalo2019}. They 
introduced a Remote Control with Digital Twin (RCDT) system, which allows humans to remotely control robots over long distances. The Digital Twin concept allows the human to have immediate feedback using a model of the robot. Based on a 5G tactile network, Miao \textit{et al.} introduce a telesurgery robot~\cite{Miao2018}. They further propose the optimisation scheme of aspects related to Human-Machine Interaction, considering Edge-Cloud Integration, Network slice, and Intelligent
Edge-Cloud. 
5G is however still far from being largely available worldwide and especially in developing countries its introduction will likely need a few more years.
Looking at the current robotics software trends, the Robot Operating System is more and more the predominant player as middleware, able to make several different modules (nodes) to talk with each others, with a very active and increasingly enthusiastic community. Therefore, approaches for remote robotics using the ROS infrastructure have started to be studied. Hajjaj and Sahari presented a Port Forwarding approach, in order to connect multiple robotic systems together over the Internet~\cite{ROS-PW}. The paper involved several stages of setup and configuration, including contacting Internet Service providers, router configuration, and configuration of the devices themselves. There was also a detailed workaround described for working with dynamic IP addresses, involving hosting a configuration file where IP addresses for each component of the system were updated upon change. However, this method had several drawbacks. 
To successfully establish a port forwarding architecture requires contacting ISPs, configuring routers with administrative privileges, and configuring the robotic system itself as well. Moreover, this method does not cater for dynamic IPs. Many routers are configured to assigned a random internal IP address to a device upon connection. Such a case requires the router configuration to be re-done, as well as the configuration on the robot itself to allow incoming communications on specific port ranges. Especially in developing countries, some of the necessary features which must be set up in order to use port forwarding are not available. Many ISPs do not enable port forwarding by default. The additional overhead of having to download an updated configuration file every time an IP address changes may cause unintended problems, and may slow down smaller, embedded systems. Port forwarding also prevents multiple devices from being connected to the same router since a specific port range is allocated to one device, and many times, the robotics applications and services use random ports and it is usually very difficult, if not impossible, to specify which range of ports should be used for communication on a case by case basis. This motivated us to start looking for computer networking methods which do not require port forwarding, can handle dynamic IP addressing, support transmission of large amounts of data efficiently, have high reliability and availability comparable to network setups with port forwarding, or have a mixture of as many of these characteristics as possible. 


\section{Network Approaches}
\label{s:na}
In our quest to answer our research question, we sought to investigate networking methods that were less cumbersome than port forwarding, but still reliable, and able to handle dynamic IP addressing, and eliminate the need for port forwarding or changing ISPs entirely, if possible. 
    We analysed the use of Mesh Virtual Networks, VPN Servers and Proxy Servers.
\subsection{Mesh Networks}
Mesh networks are a type of network topology in which each host or node in the network is able to connect with any and every other host in a dynamic, non-hierarchical manner. This contrasts with the tree-like protocols such as the Spanning tree protocol, where each node is connected to a small subset of the nodes in the network. Mesh networks (also called meshnets) have the ability to dynamically self-organize and self-configure, which is advantageous in events such as the failure of multiple nodes in the network. Thus they are capable of increased fault tolerance, and require less supervision and maintenance. Many meshnets employ high degrees of encryption, ensuring that data moves from source to intended target and is not mis-routed or accessible by unpermitted users on the network. We considered two mesh networks as alternatives to Port forwarding: CJDNS and Yggdrasil. 

CJDNS~\cite{cjdns}, originally developed by Caleb James DeLisle and written in Node.js and C, implements an encrypted IPv6 network using public-key cryptography for address allocation and a distributed hash table for routing. This provides near-zero-configuration networking, and prevents many of the security and scalability issues that plague existing networks.

On the other hand, Yggdrasil~\cite{yggdrasil}, community-developed and written in Go, is an early-stage implementation of a fully end-to-end encrypted IPv6 network. It is lightweight, self-arranging, supported on multiple platforms and allows pretty much any IPv6-capable application to communicate securely with other Yggdrasil nodes. Yggdrasil does not require you to have IPv6 Internet connectivity - it also works over IPv4. Although Yggdrasil shares many similarities with CJDNS, it employs a different routing algorithm based on a globally-agreed spanning tree and greedy routing in a metric space, and aims to implement some novel local backpressure routing techniques. 

We shortlisted these meshnets because of the ability to connect as many hosts as needed, and of the fact that communication was fully end to end encrypted. The handling of dynamic IP addresses was still to be further investigated. However, from their documentation, they appeared to need some additional post-installation configuration, which appeared slightly technical. This may discourage less advanced robotics systems users from considering using it as a networking approach. There was also not much documentation on whether port forwarding was required or not, leaving this to be verified.
\subsection{Husarnet (P2P VPN With Base Server)}
A VPN enables a host to connect to the internet anonymously, typically by relaying the traffic through an intermediary node. Among the various VPN-based solutions, we chose to explore Husarnet~\cite{husarnet}, because it employs an approach similar to ROS: a central node (or cluster of nodes) is responsible for connecting the individual nodes on the network with each other and exchanging address information between them; after this is successfully completed, the information exchanged between the nodes passes only between the nodes themselves, i.e. a peer-to-peer connection. It also shares some similarities with the Mesh networks described earlier, i.e. using IPv6 addressing, etc. but does not allow every node to freely connect to every other node. It creates virtual networks for each individual user, and the devices connected to this virtual network are then freely allowed to establish p2p connections with each other. It is also less technical and more user friendly than its meshnet counterparts, requiring minimal configuration on the user side, and integrating a web application to aid in management of the individual virtual networks.
Husarnet was also selected with the viewpoint that, in a large number of robotic systems, only a few nodes need to be connected in order to exchange information. The feature of being able to create isolated virtual networks capable of peer to peer connections offered possibilities for comparison to port forwarding. However, there was a risk, that if ever the Base servers were to go offline, the network as a whole, including the individual virtual networks would not function as they should, resulting in the nodes not being able to establish connections to each other. This could lead to undesired unavailability of the network.
\subsection{Remote.it (Proxy Server)}
Remote.it~\cite{remote.it} is a software package developed by remot3.it Inc. It is a tool that enables devices to communicate with one another by interacting with a middle-man server cluster that provides a random port linked to the requested port on the remote host, and forwards all communication to and from the remote host through only that specific port. The ports are linked to services or applications, such as SSH, VNC, and game servers. It boasts a large library of mobile, web and desktop applications available to the user, which enable greater ease of access. It also has an extensively documented API which allows developers to create custom applications integrating their software. \\
Though much closer to the idea of ports rather than IP addresses, remote.it was chosen for investigation as well, since it offered the chance to determine which specific port would be in use, and offered individual access to  the resources connected to those ports, thus eliminating the need for dealing with public IP addresses as opposed to port forwarding. 
However, since the service requires and gives access to individual ports on the remote hosts, it is unclear what would happen in the case of using robotics applications that would require access to multiple ports simultaneously. There is also a similar risk as with Husarnet, where if the middle-man server goes down, so does all communication with the remote system, which is undesirable. 

\section{Networking Test Scenarios}
\label{s:nts}
This section will present the different types of tests chosen, with the motivation behind them. We have selected:
\begin{itemize}
    \item \textbf{Ping test}:
    we aim to measure the round trip times of packets sent to robotics systems in different locations with these networking approaches. We wanted to investigate whether the geographical distance and the technique used affected the round trip times. For this test, we executed a shell script that used the ping command to obtain roundtrip time data and parse it into CSV format;
    \item \textbf{iperf tests (TCP + UDP)}: TCP and UDP are tested since they are two of the main protocols used to exchange information between two hosts. UDP is used in contexts such as image streams, where confirmation of the receipt of the packages is generally not necessary, whereas TCP relies on the acknowledgement of the successful receipt of packets. ROS relies heavily on TCP, therefore we use iperf to check if there would be a difference in transmission speed depending on the protocol, and also to compare TCP with ROS transmission speeds. We used the available option of iperf to output data in csv format.
    \item \textbf{SSH speed test}: a lot of remote work is done over SSH, therefore we need to ascertain the speeds achievable on each of the networking methods over SSH. We also wanted to see if the speeds were comparable to ROS transmission speeds. We use a shell script that creates a file of size 10MB and uploads it to the remote host, then downloads the same file from the remote host, all while measuring the transmission speeds.
    \item \textbf{ROS Node Image stream test}: we want to evaluate the transmission of an image stream over ROS, because of the popularity of ROS in robotic systems. This is an essential part of tele-operation. We use a ROS node that streams images using OpenCV, the output of the \texttt{rostopic bw} command to calculate the bandwidth needed, which we converted to CSV by means of a helper Python 3 script.

\end{itemize}

The tests were automated using several Bash scripts which utilised command line utilities such as \verb|ping|, \verb|iperf|, \verb|ssh| and the ROS family of commands, coupled with a Python3 script to plot the data from the generated CSV files.  
We structured the tests such that we would perform them on multiple days at random over a four month period in order to reduce the bias coming from specific network conditions on a specific day. The overall data was then analysed together. The tests runs were repeated at least five times. All tests except the ssh speed test were ran for 30 seconds each. Since the ssh test relied on transmitting a file of a specific size, the completion time varied according to the network speed. 

\section{Experimental results}
\label{s:er}
\begin{figure*}
\includegraphics[width=.5\columnwidth]{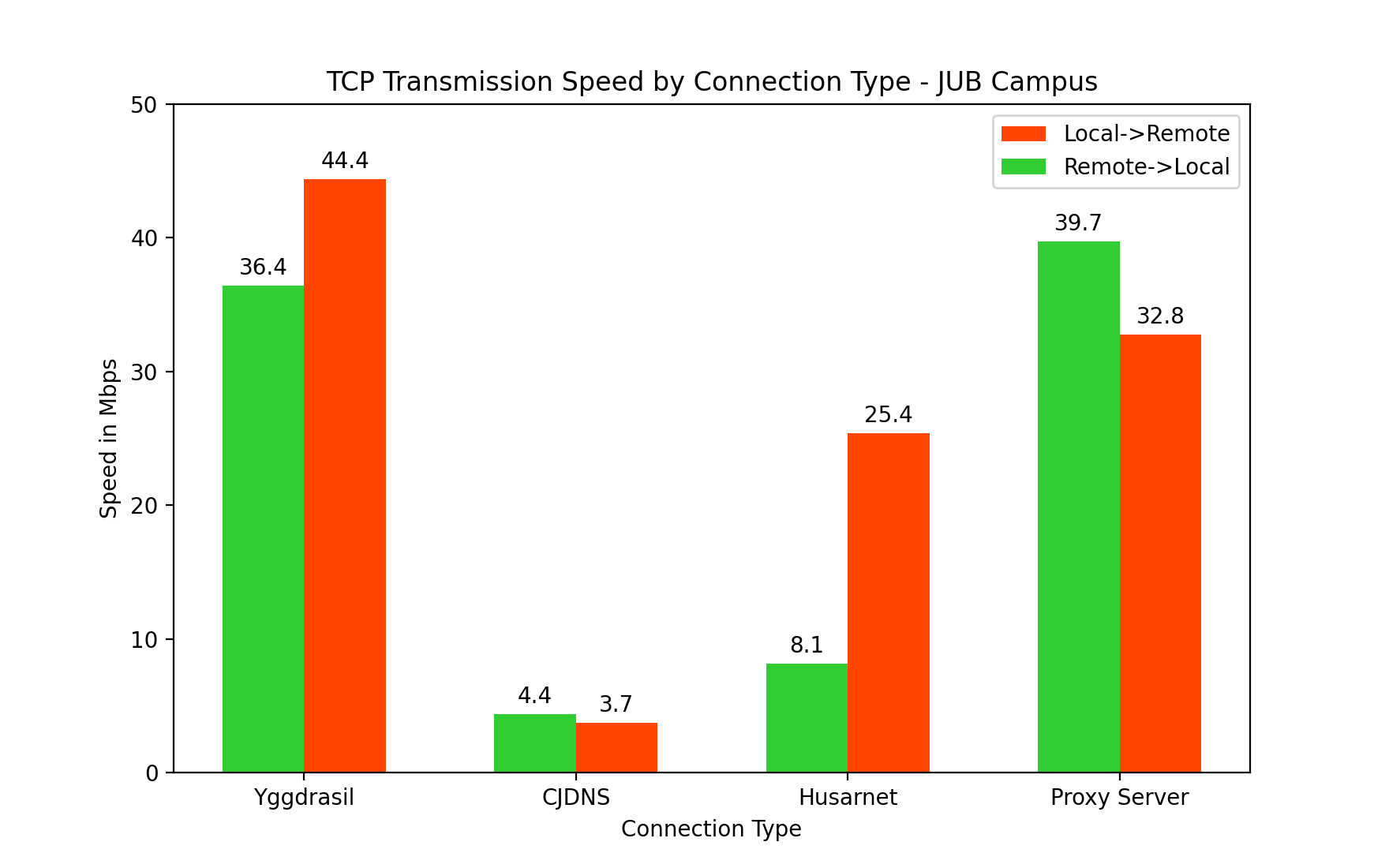}
\includegraphics[width=.5\columnwidth]{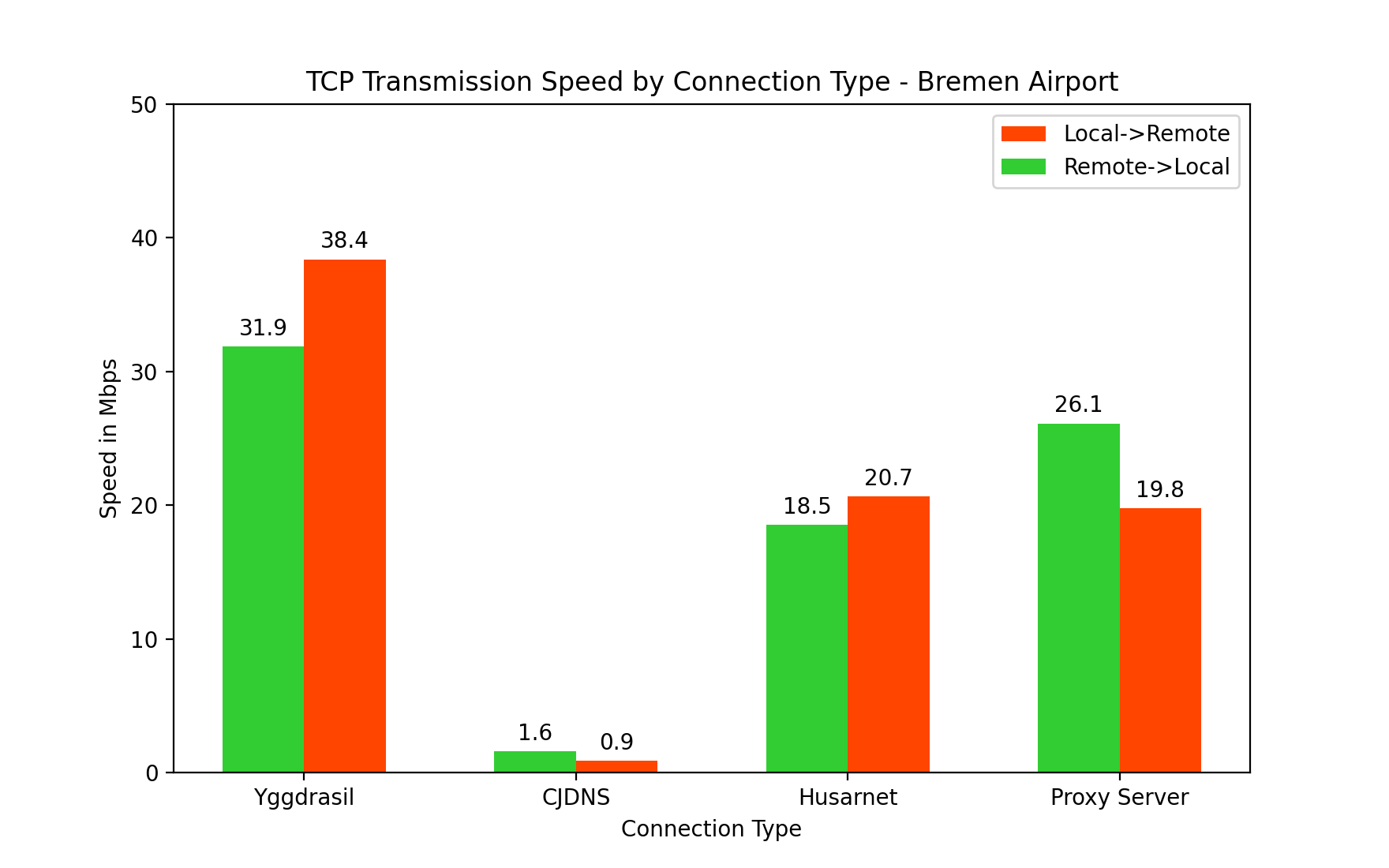} 
\includegraphics[width=.5\columnwidth]{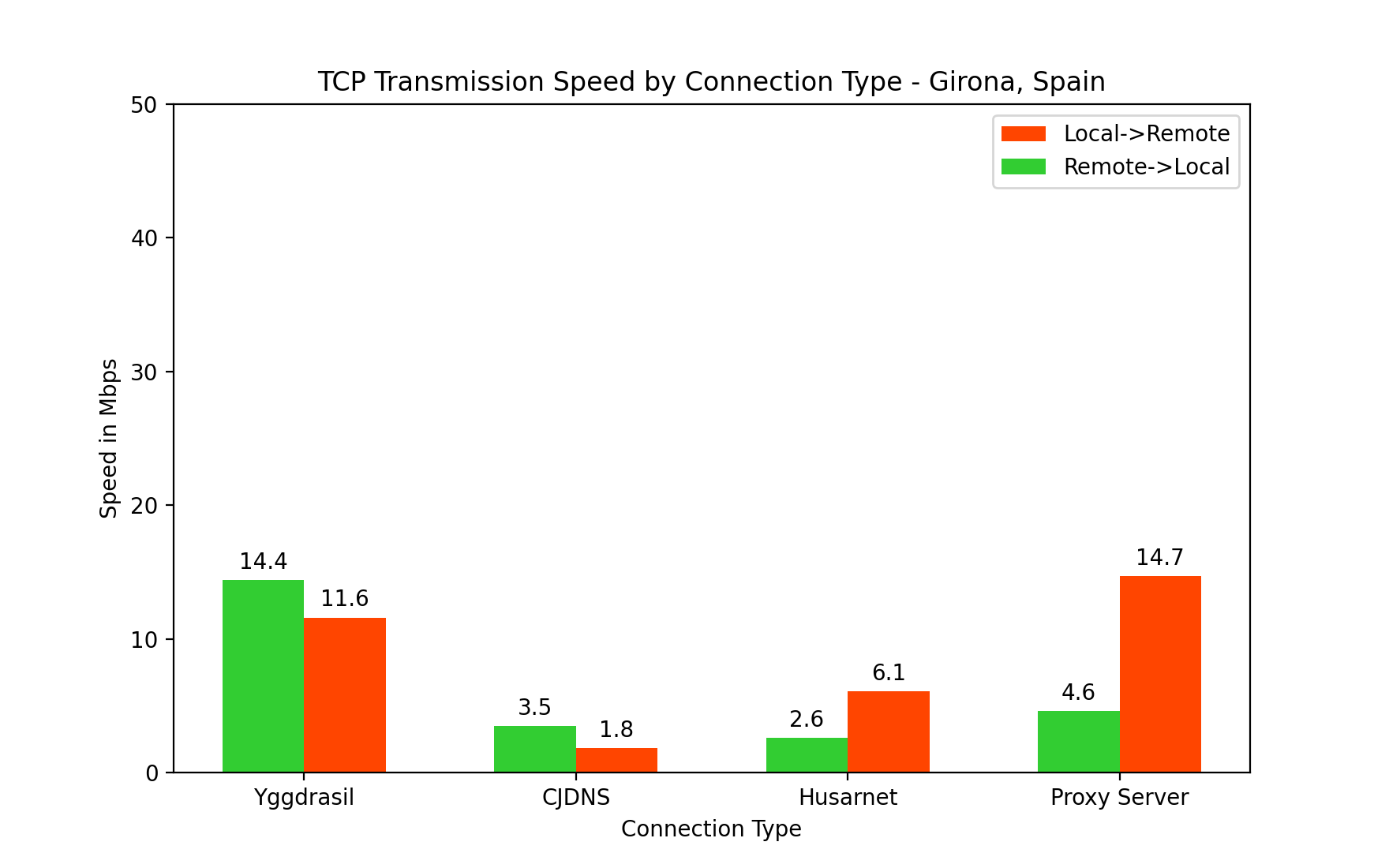}
\includegraphics[width=.5\columnwidth]{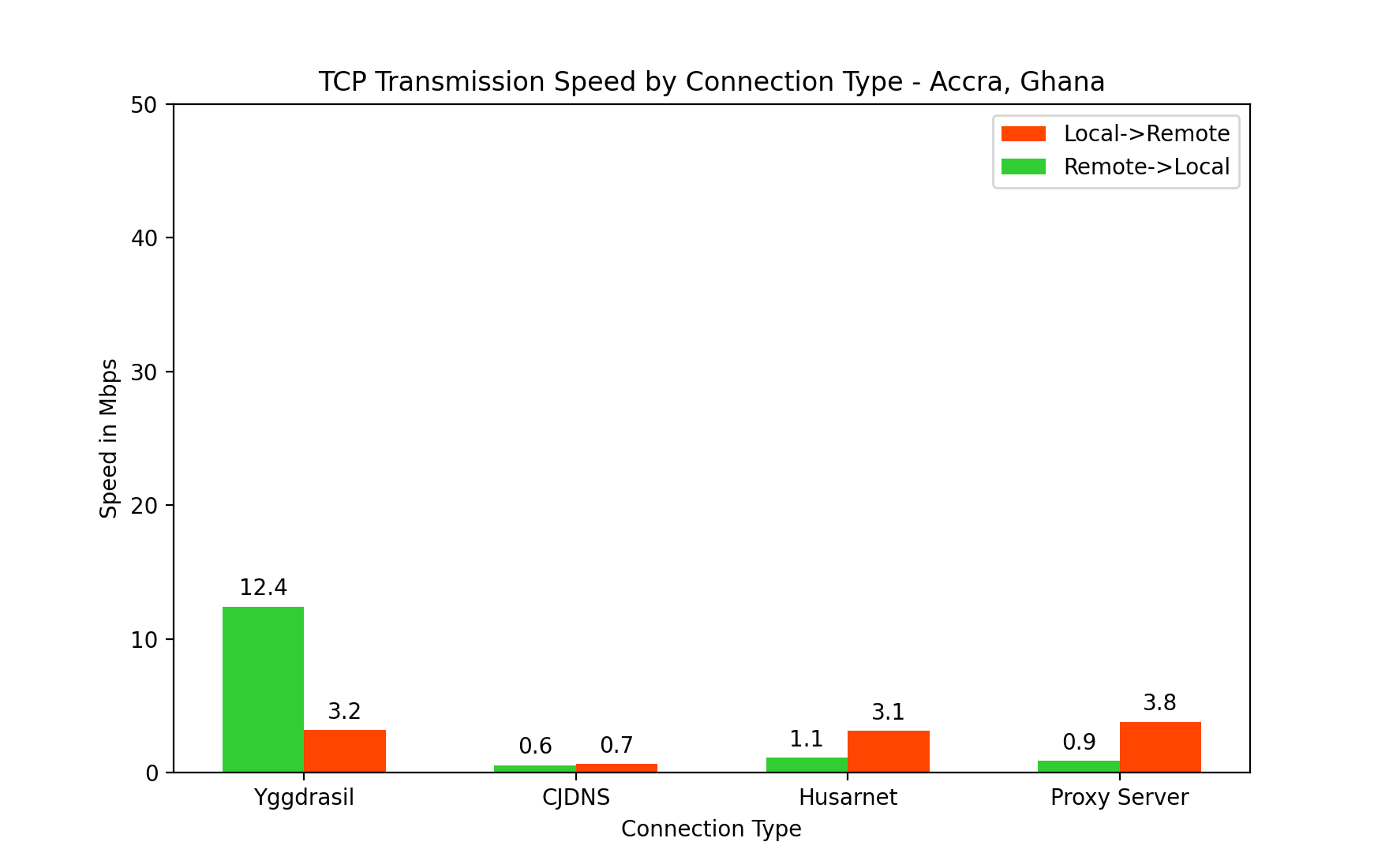}
\centering 
\caption{Bar chart plots of TCP transmission speed tests between different locations using iperf - local, urban, Germany-Spain, Germany-Ghana}
\label{iperf_tcp}
\end{figure*}

\begin{figure*}
\includegraphics[width=.5\columnwidth]{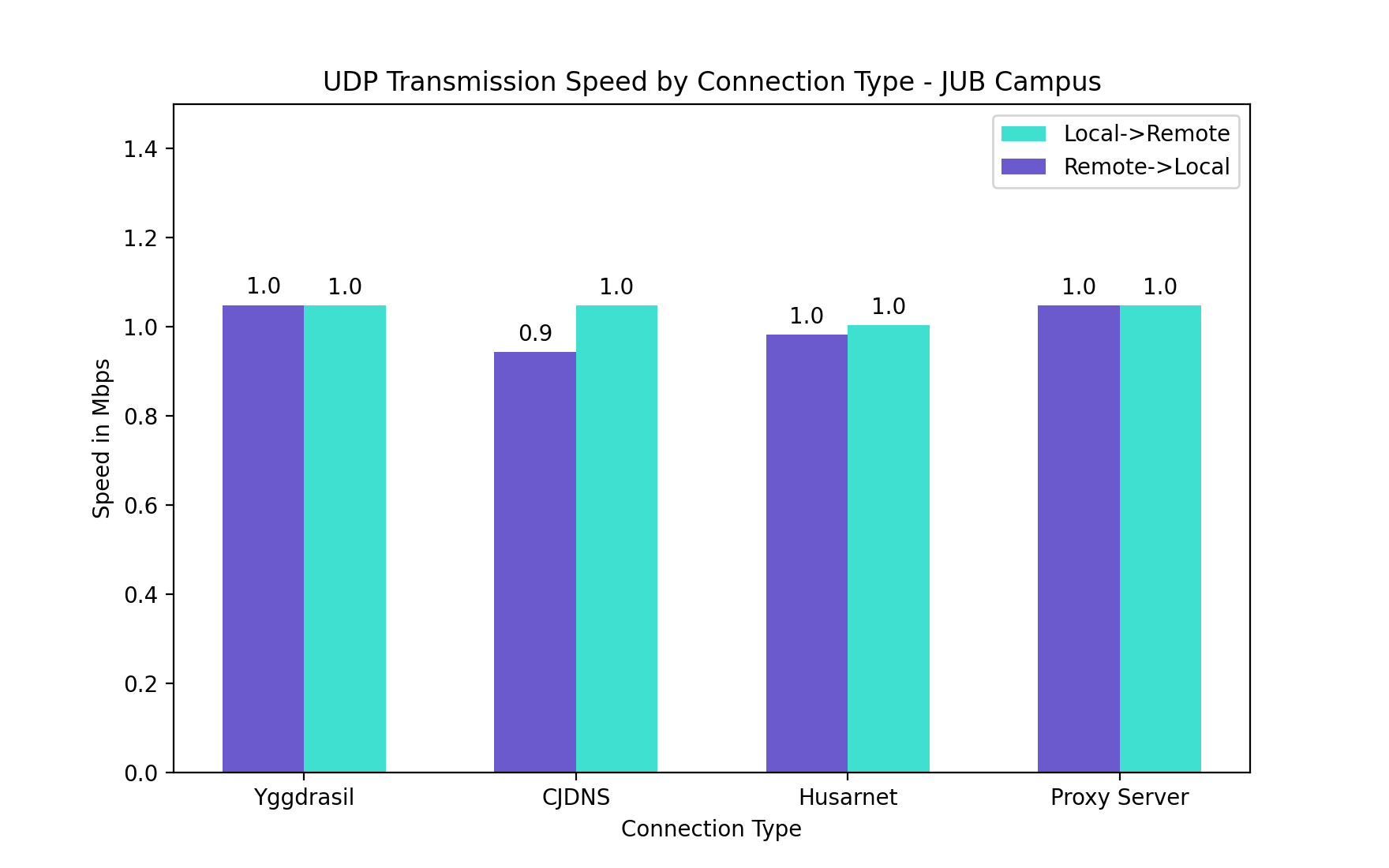}
\includegraphics[width=.5\columnwidth]{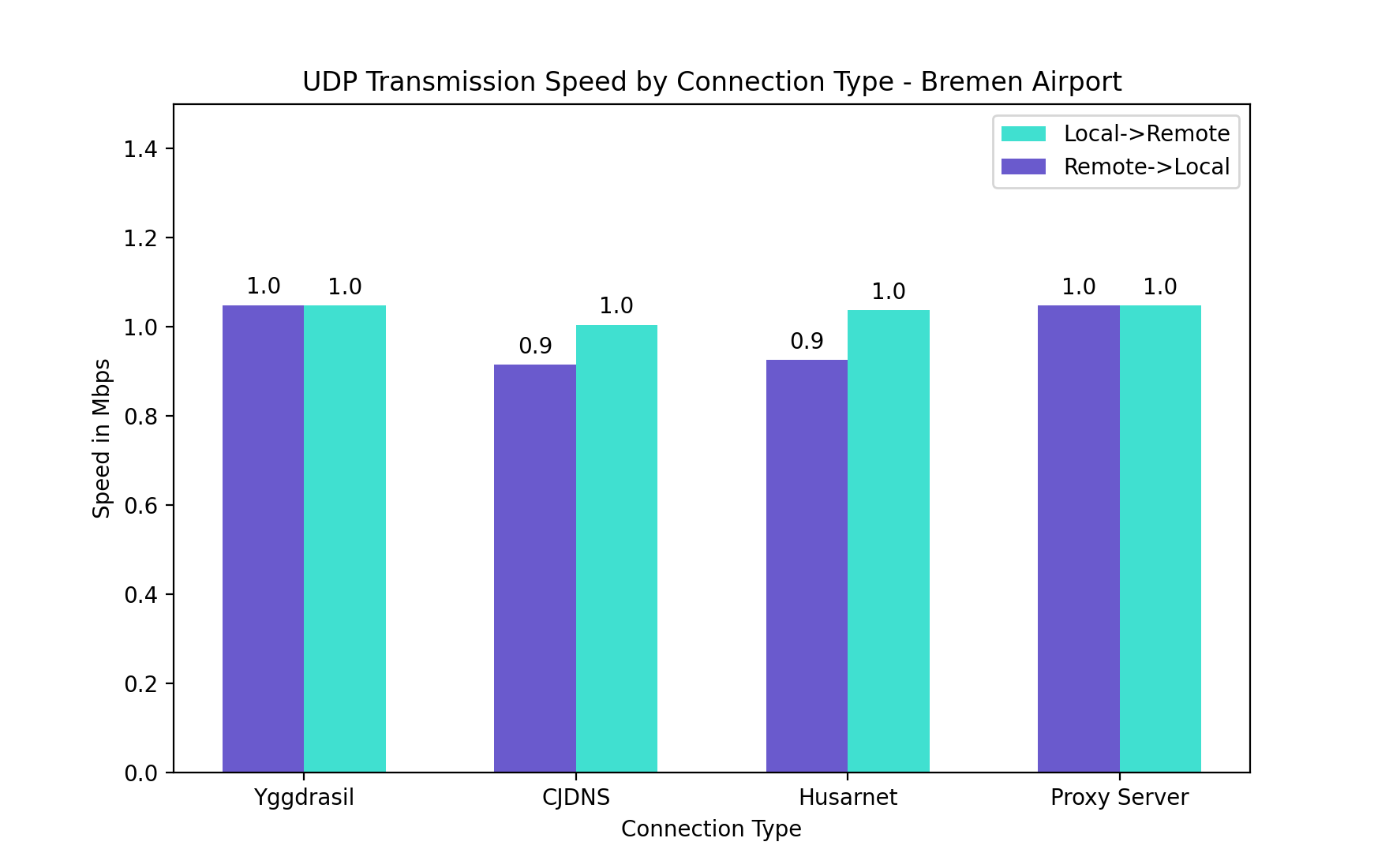} 
\includegraphics[width=.5\columnwidth]{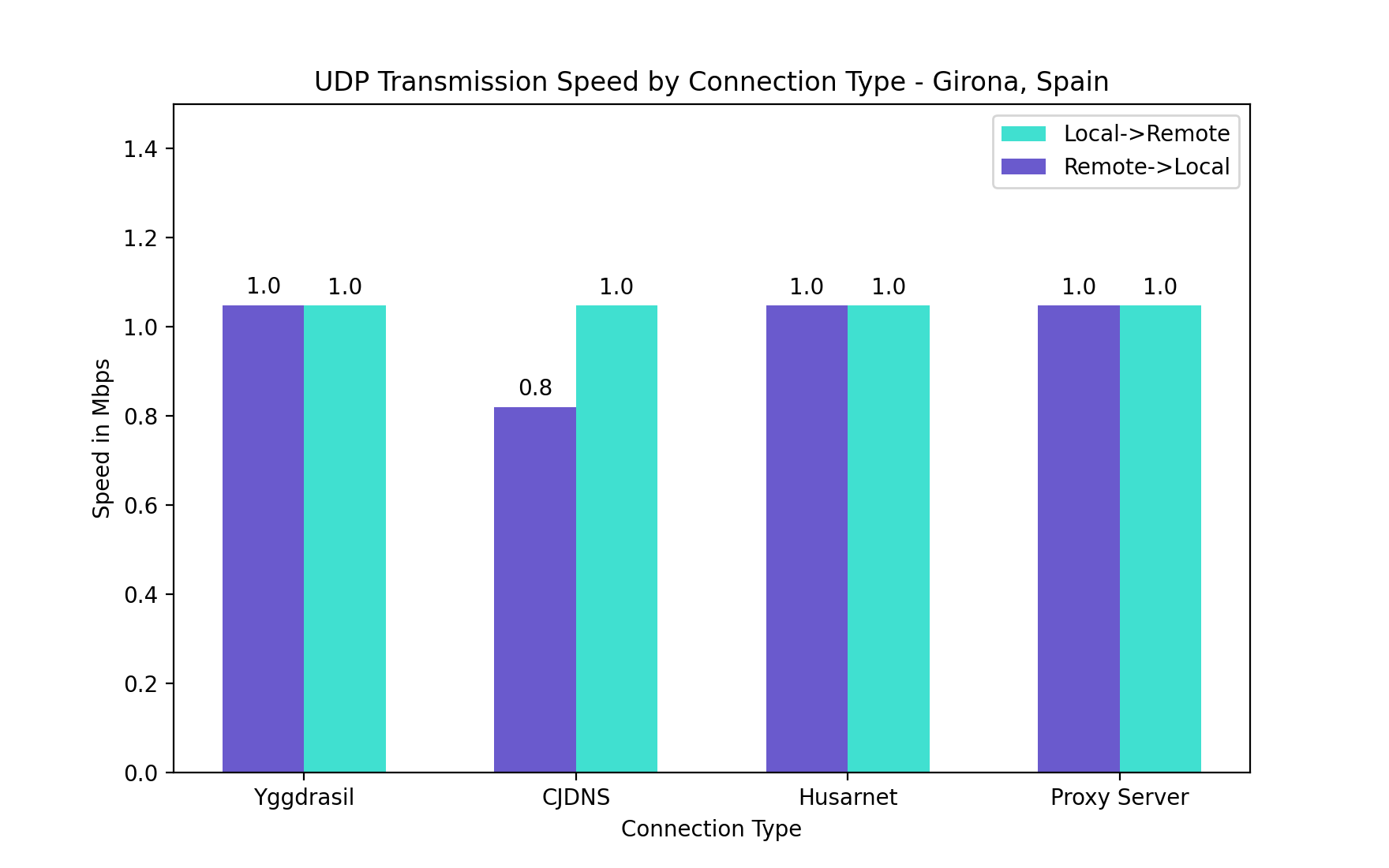}
\includegraphics[width=.5\columnwidth]{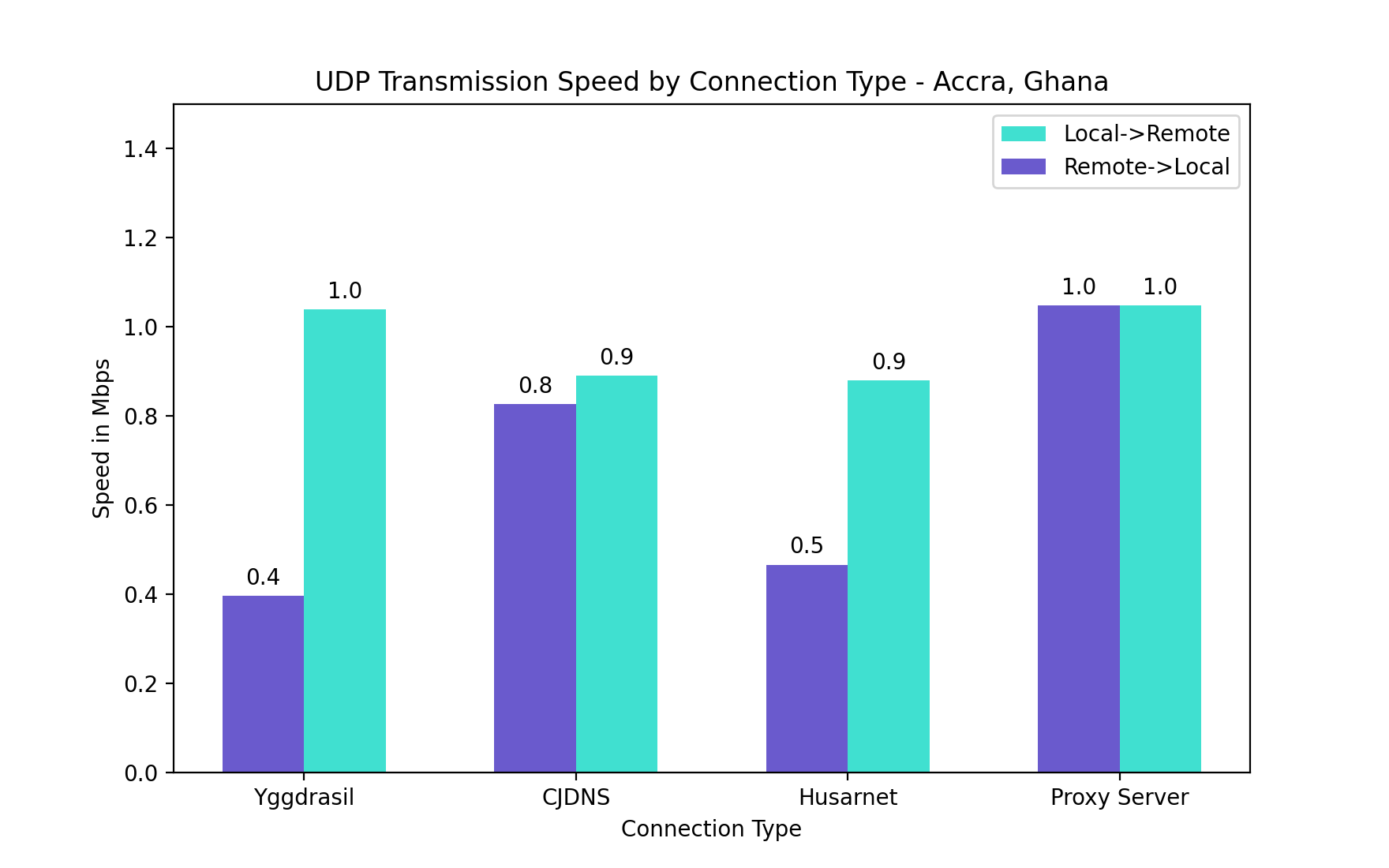}
\centering 
\caption{Bar chart plots of UDP transmission speed tests between different locations using iperf - local, urban, Germany-Spain, Germany-Ghana}
\label{iperf_udp}
\end{figure*}

\begin{figure*}
\includegraphics[width=.5\columnwidth]{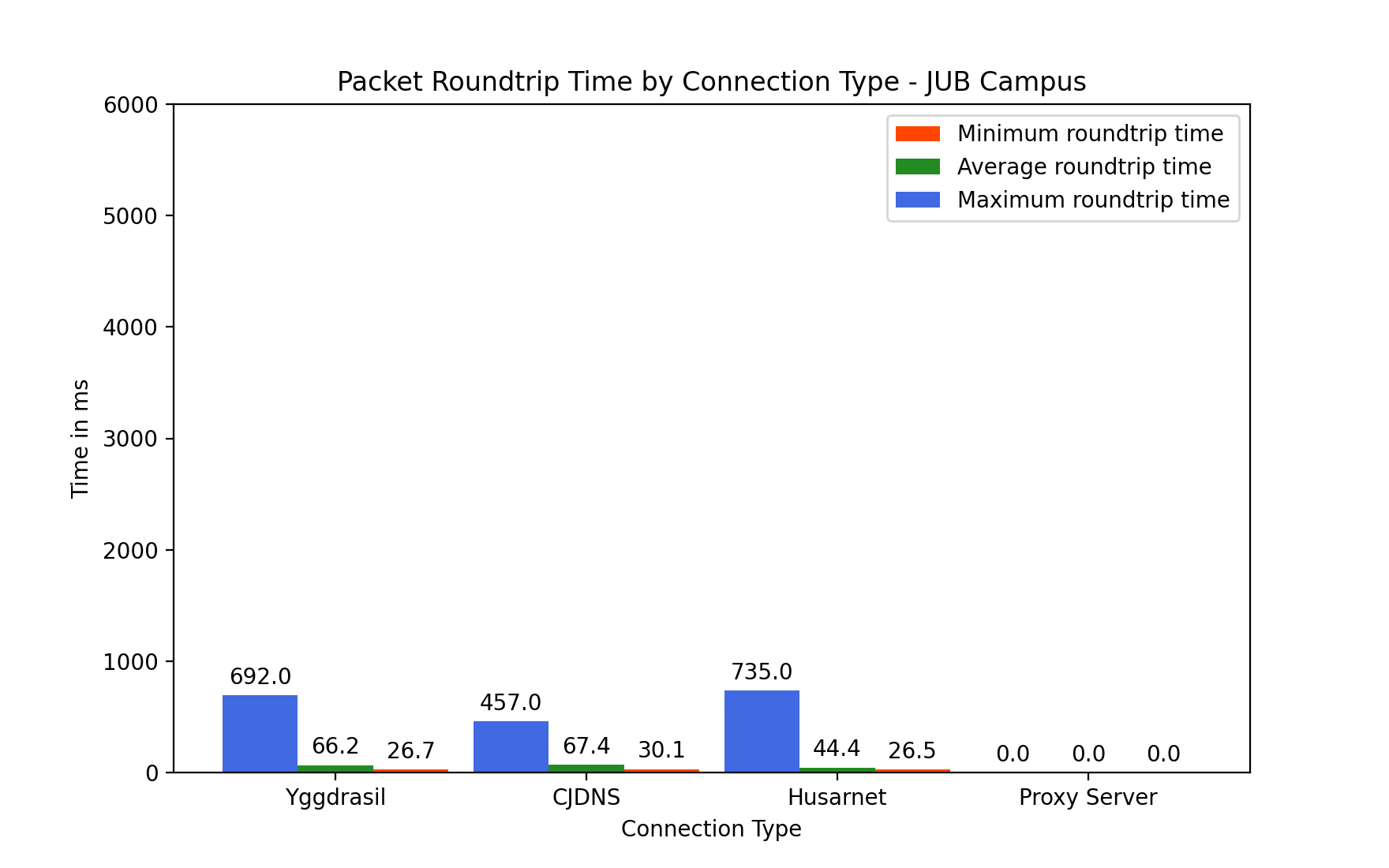}
\includegraphics[width=.5\columnwidth]{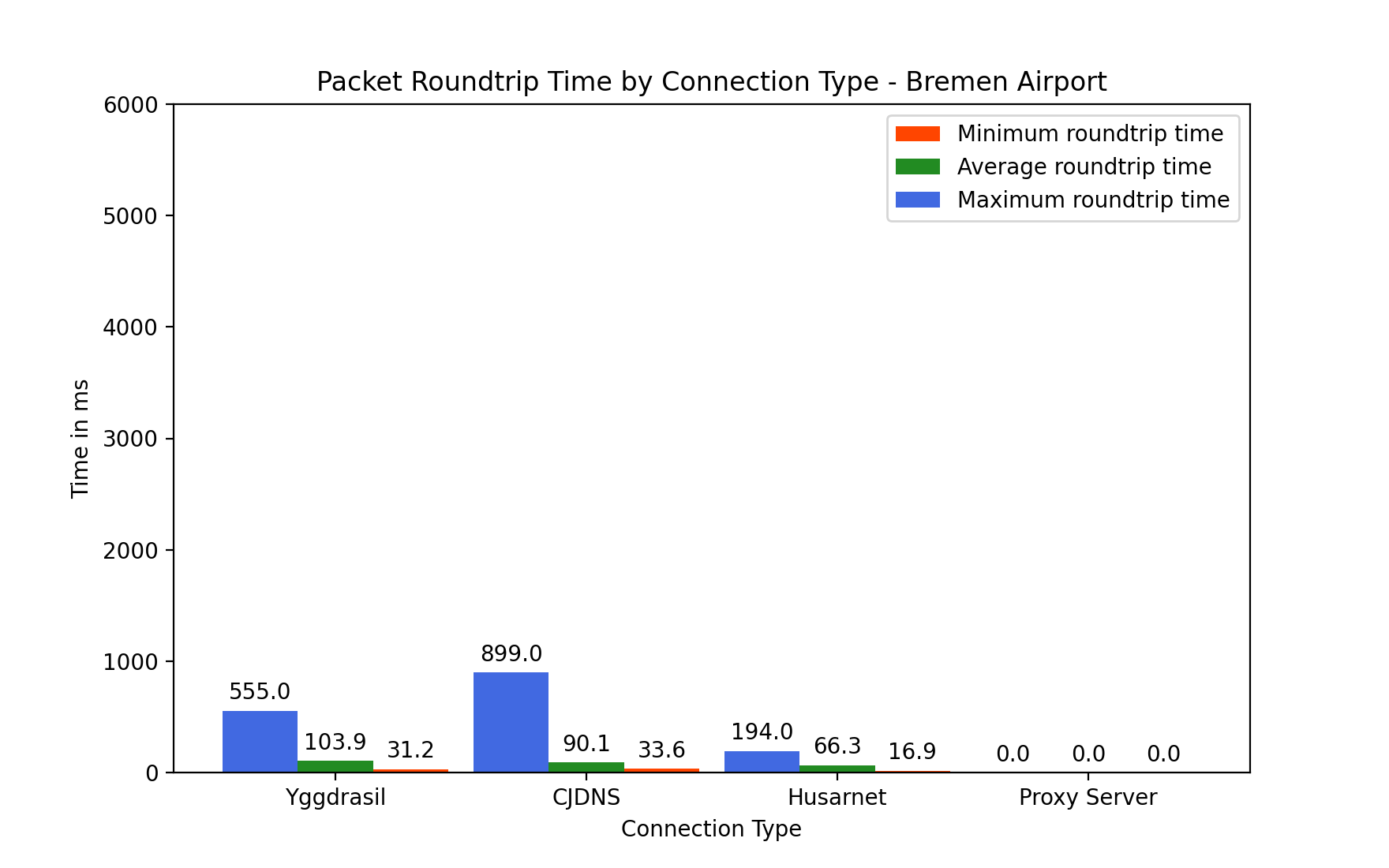} 
\includegraphics[width=.5\columnwidth]{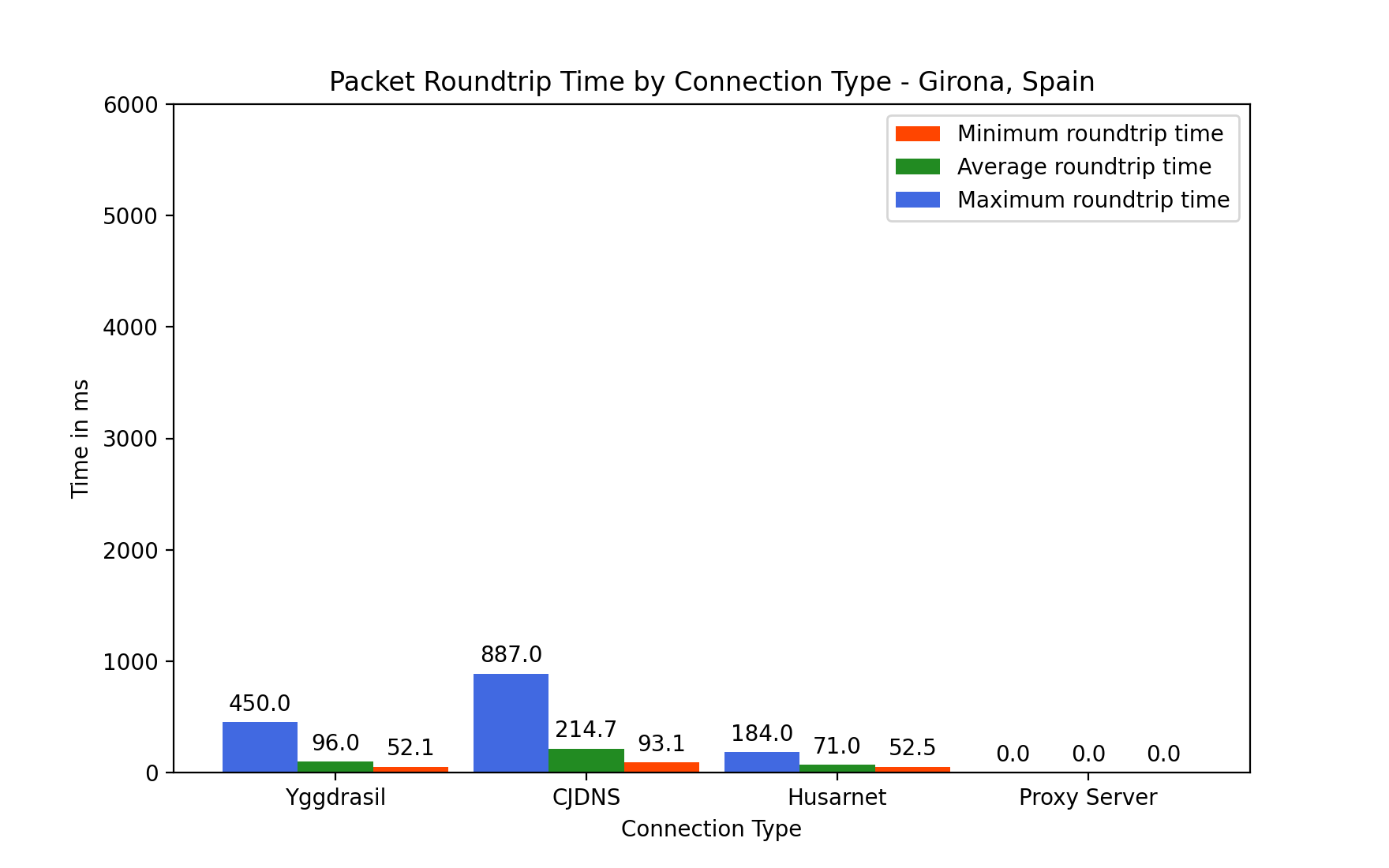}
\includegraphics[width=.5\columnwidth]{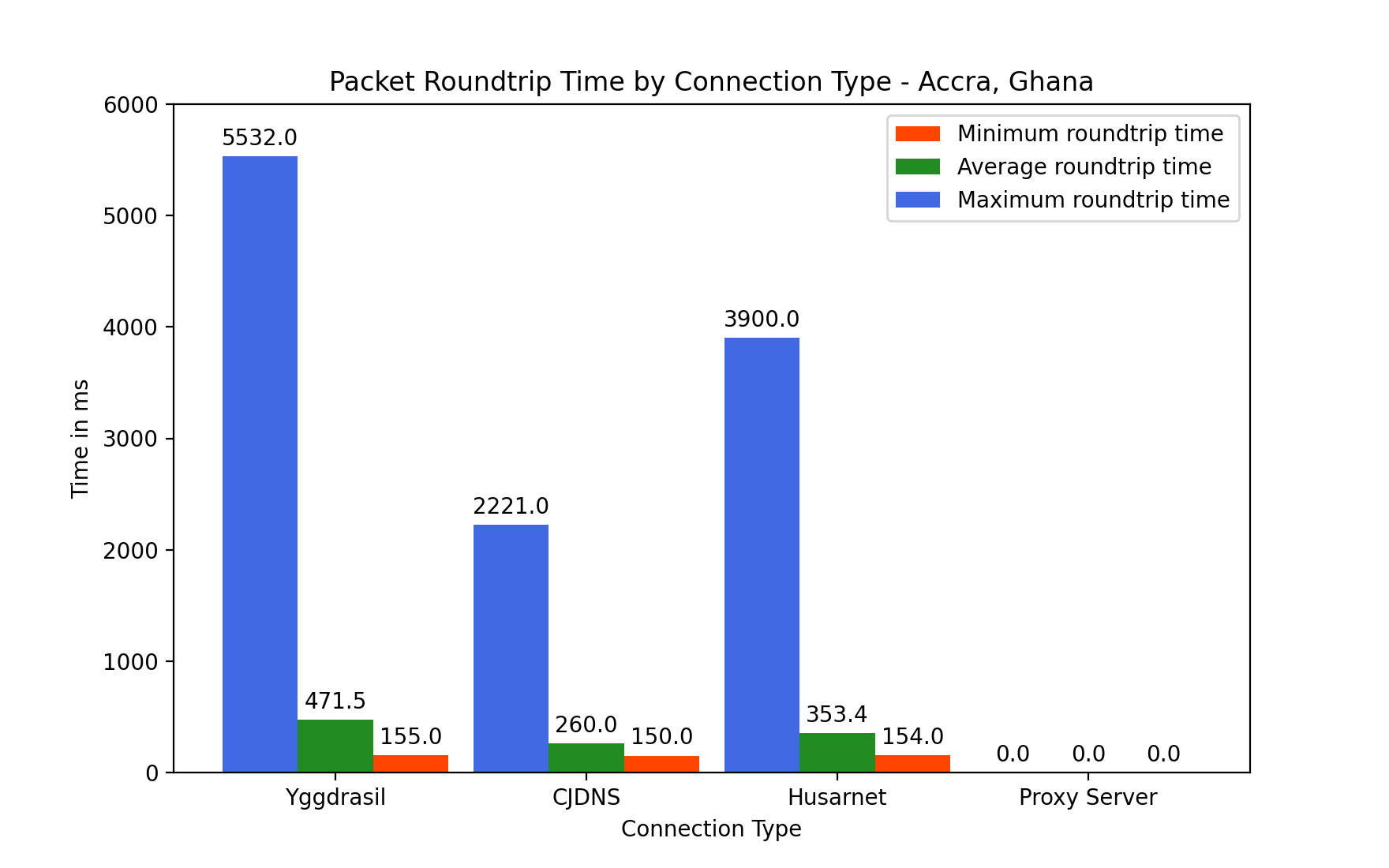}
\centering 
\caption{Bar chart plots of packet round trip time tests between different locations using ping - local, urban, Germany-Spain, Germany-Ghana}
\label{ping}
\end{figure*}

\begin{figure*}
\includegraphics[width=.5\columnwidth]{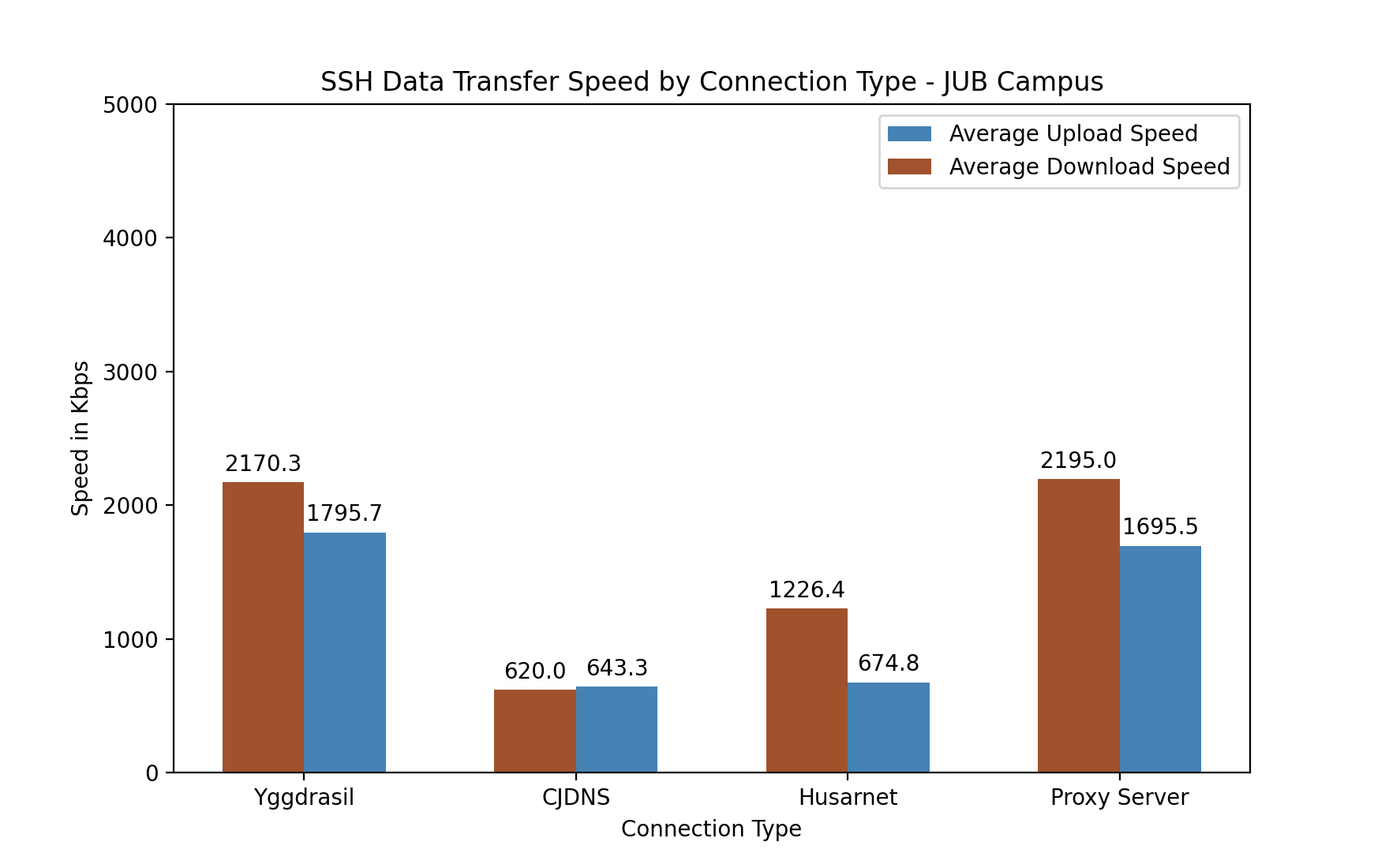}
\includegraphics[width=.5\columnwidth]{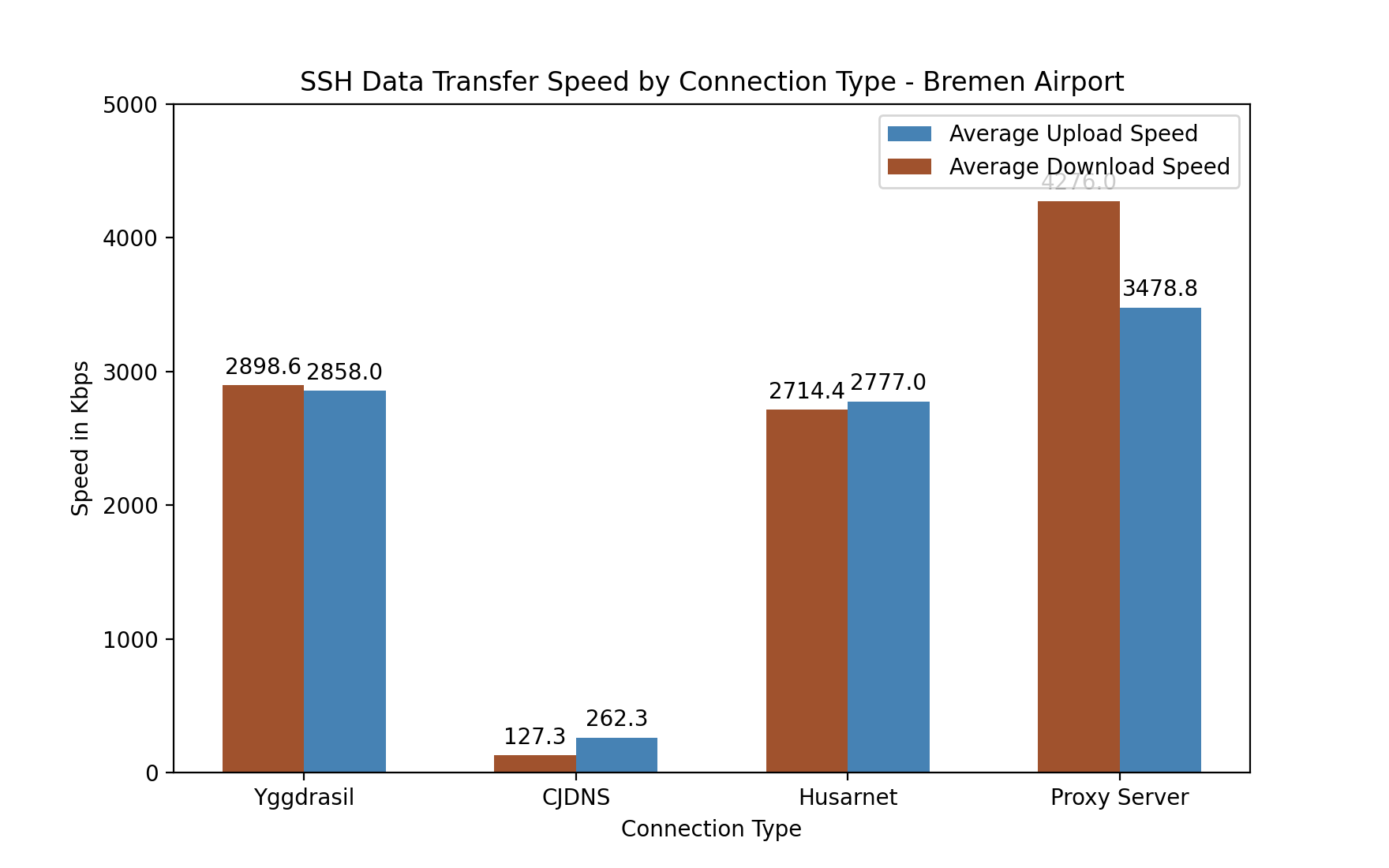} 
\includegraphics[width=.5\columnwidth]{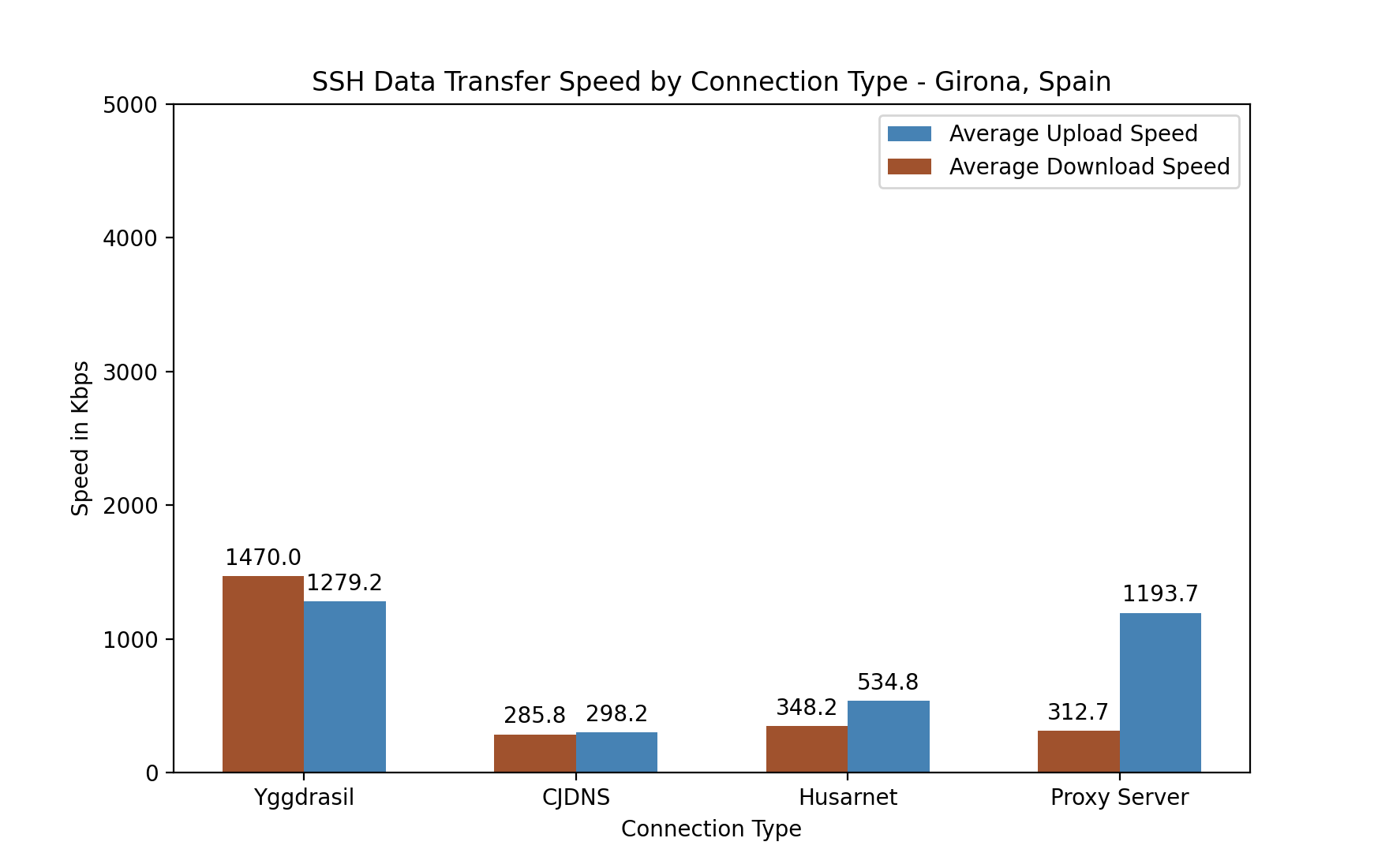}
\includegraphics[width=.5\columnwidth]{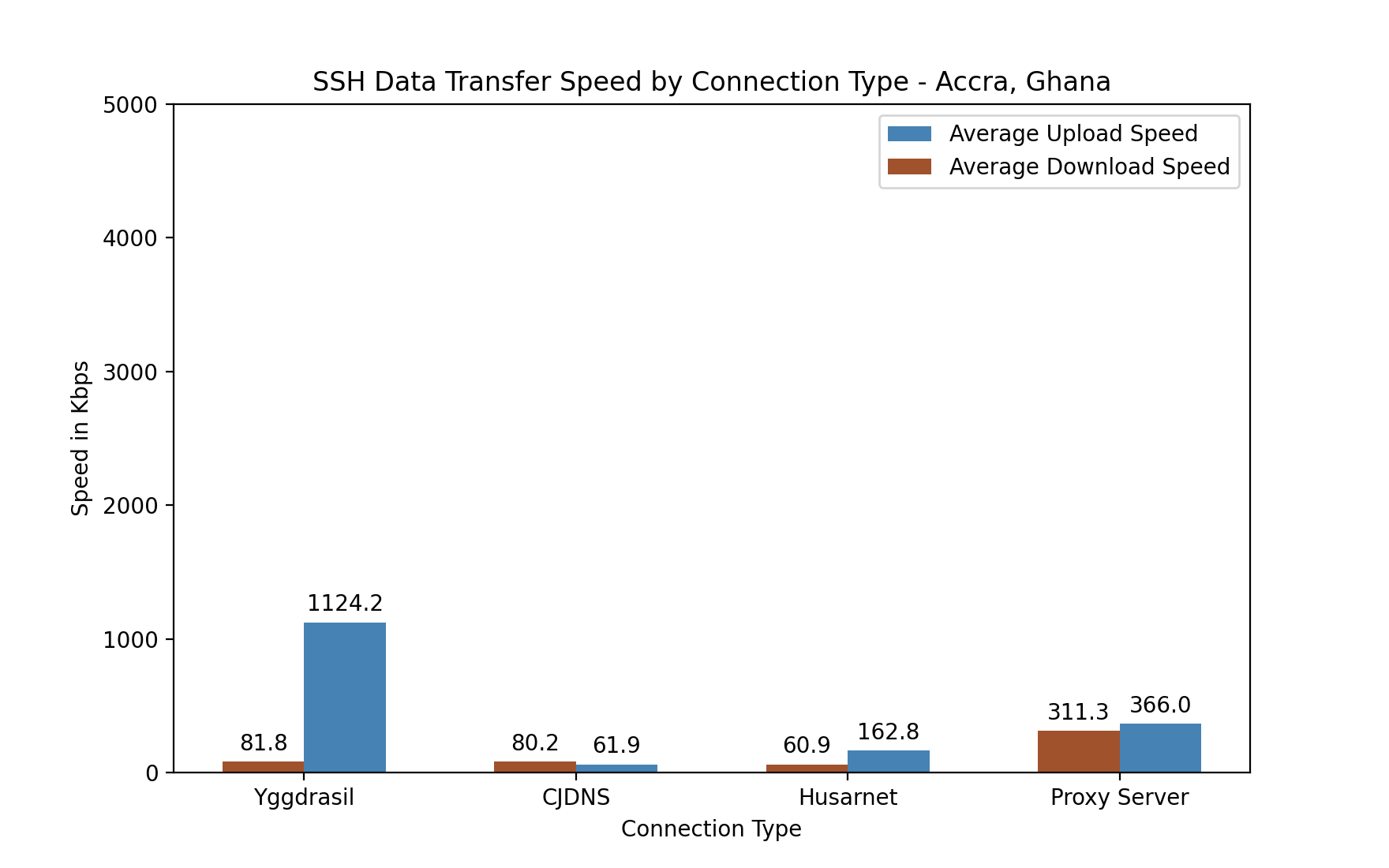}
\centering \caption{Bar chart plots of ssh transmission speed tests between different locations using scp - local, urban, Germany-Spain, Germany-Ghana}
\label{ssh}
\end{figure*}

Figures~\ref{iperf_tcp},~\ref{iperf_udp},~\ref{ping},~\ref{ssh} show the plots of the results of our experiments. The most left plots represent the results from our tests on the campus of Jacobs University Bremen, the middle left ones represent the results from tests we conducted between the campus of Jacobs University Bremen and a remote workstation near the Bremen Airport (approx. 15 km). The middle right graphs depict our results from tests between Jacobs University Bremen and a \textit{Sparus II} underwater robot located at the University of Girona, Catalonia, Spain (approx. 1500 km). Additional tests have been made between Jacobs University Bremen and a workstation in Accra, Ghana (approx. 6700 km), shown on the most right. Thus, we have 4 sets of tests: campus, city, continental and intercontinental.  The campus tests and the Bremen Airport tests were easier for us to perform since we had ready access to the systems we needed to test. However, the tests between the remaining two locations were more difficult to arrange, thus providing us with less data. 

\begin{figure*}
\includegraphics[width=.5\columnwidth]{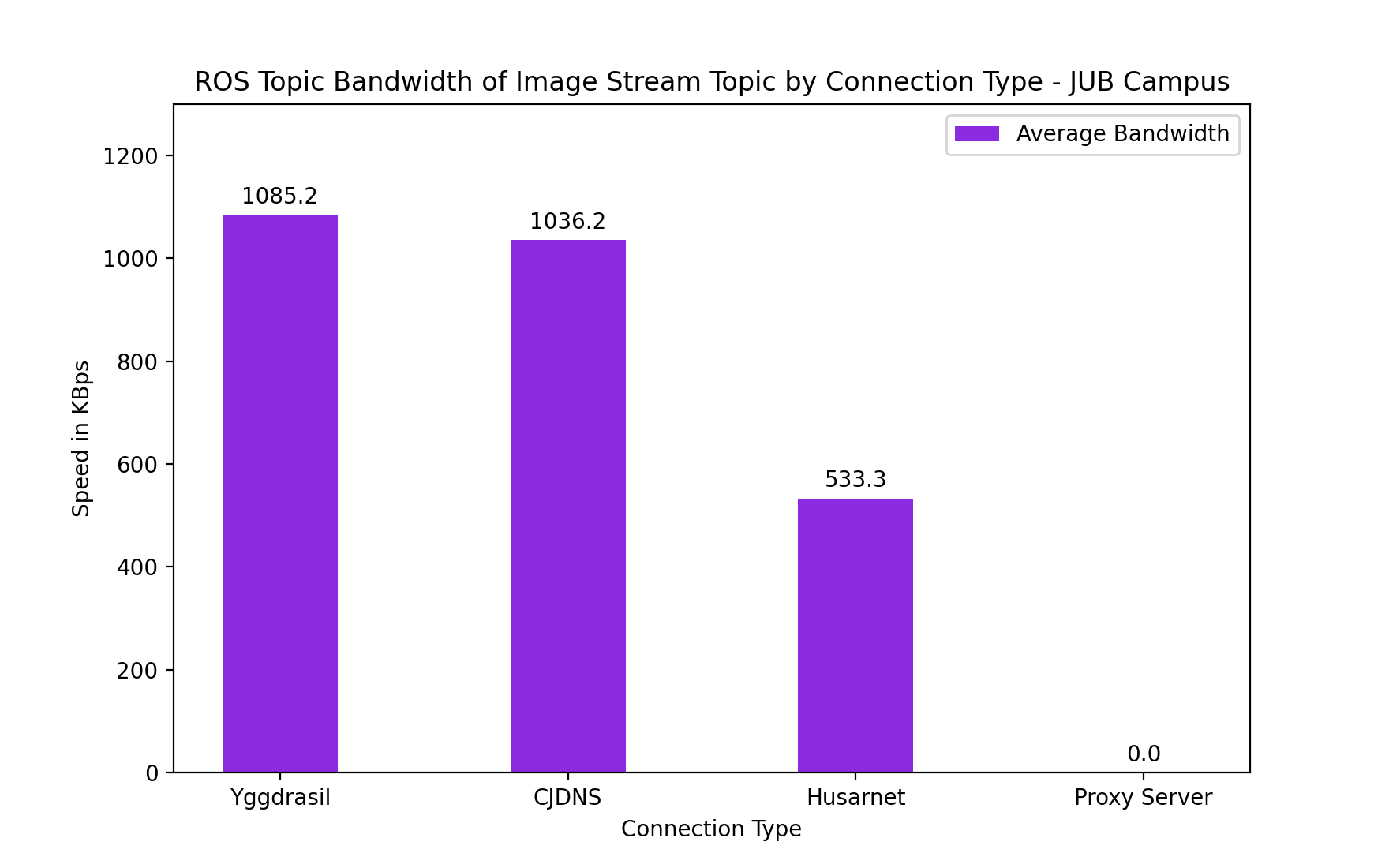}
\includegraphics[width=.5\columnwidth]{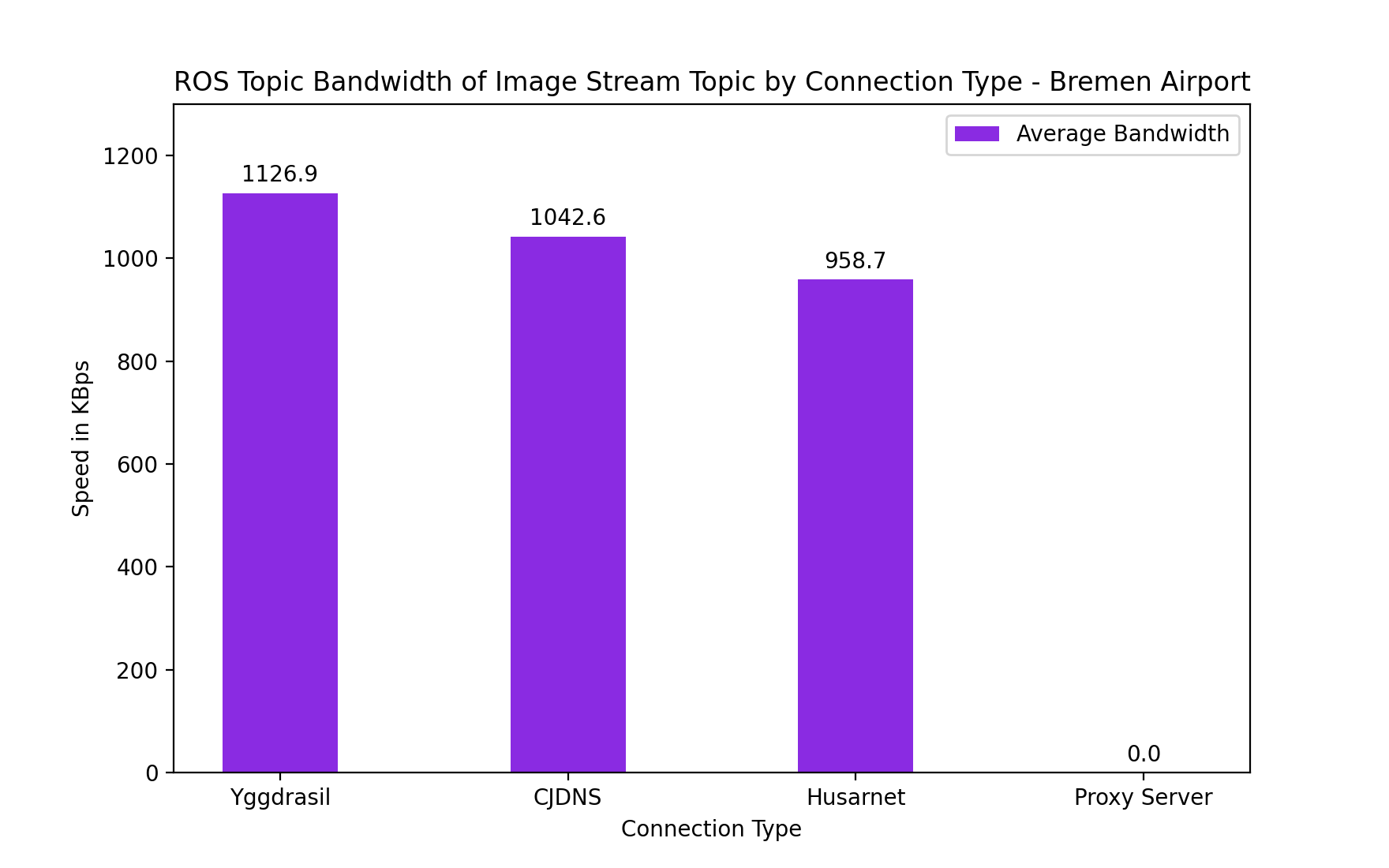} 
\includegraphics[width=.5\columnwidth]{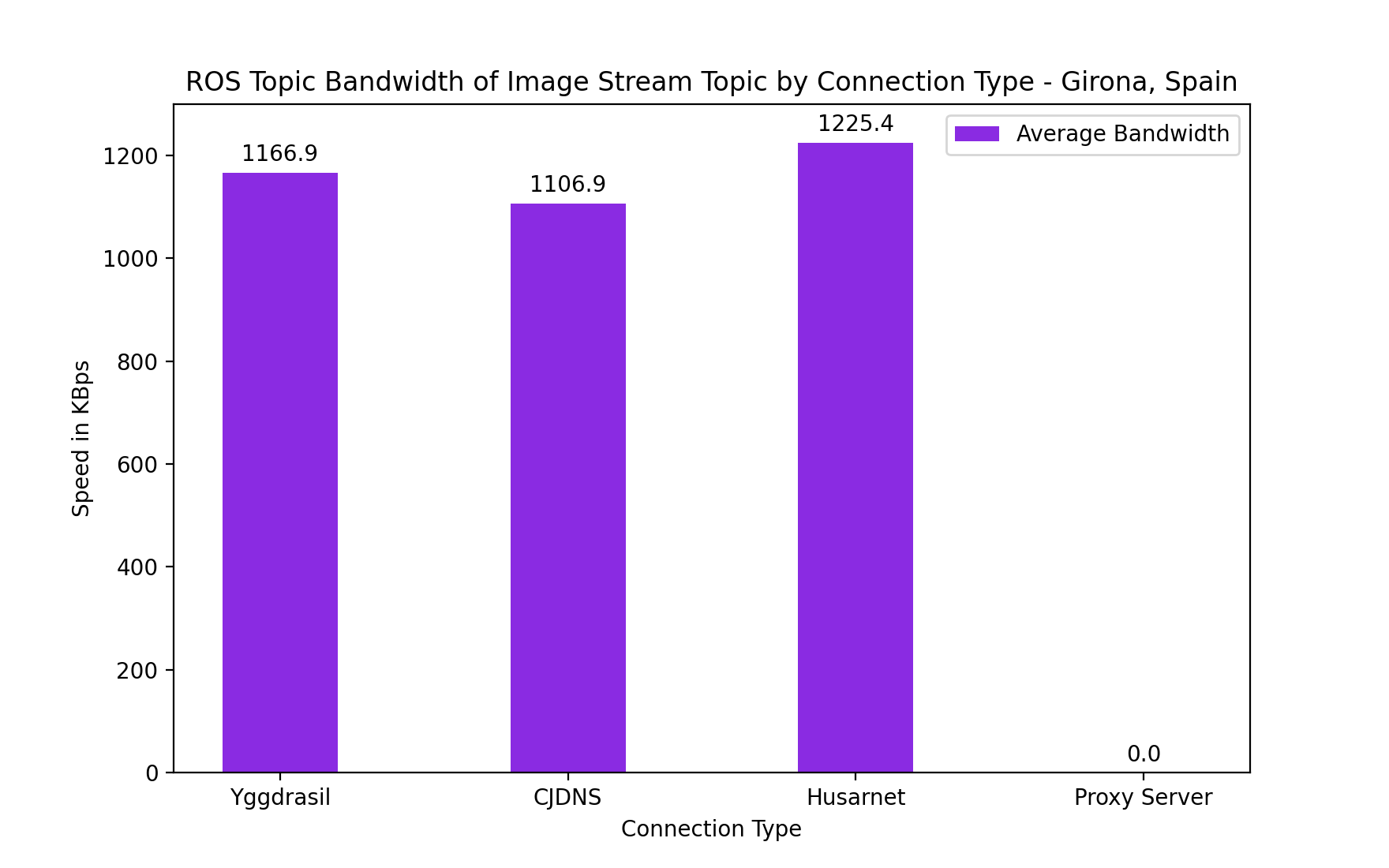}
\includegraphics[width=.5\columnwidth]{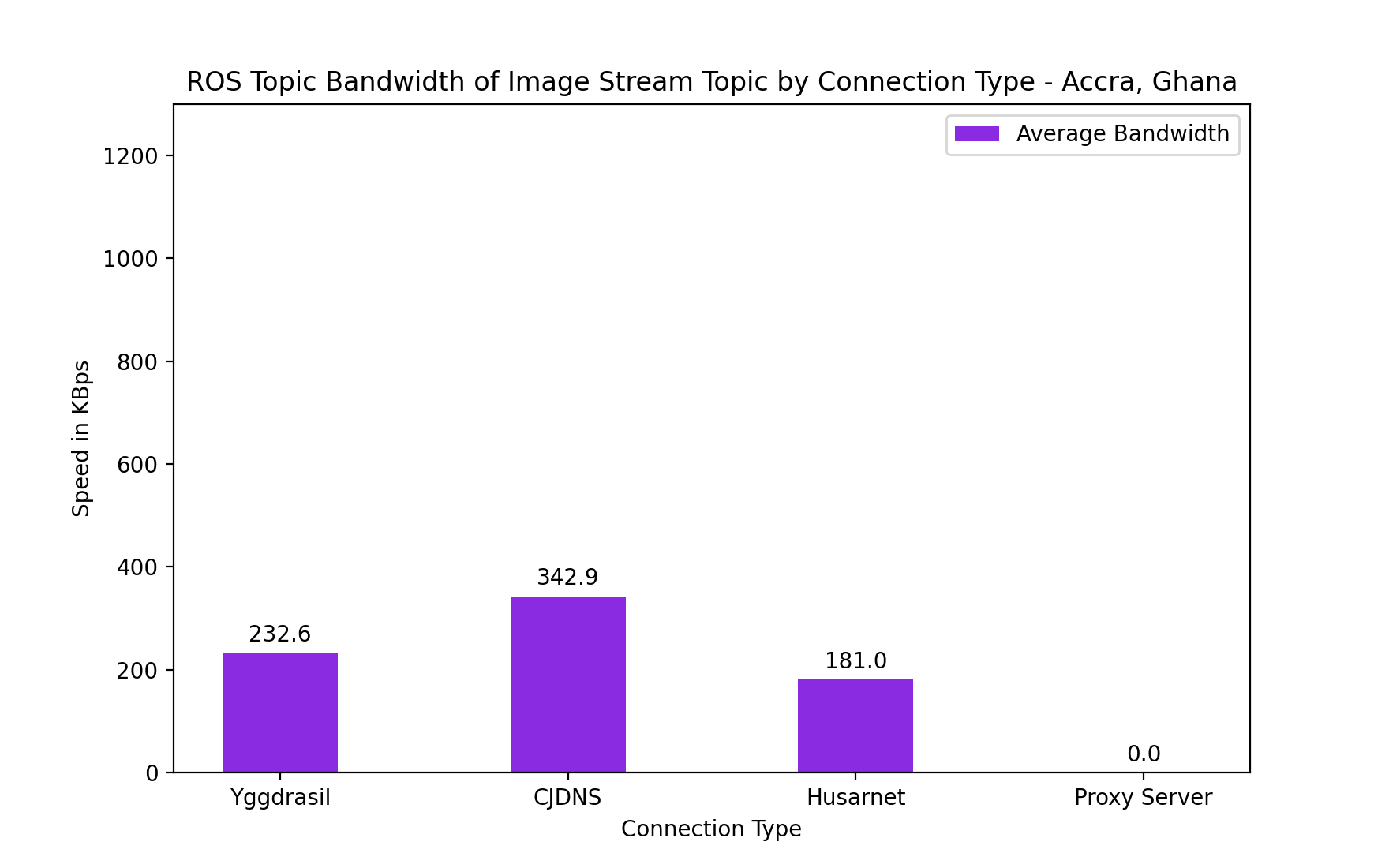}
\centering \caption{Bar chart plots of bandwidth tests of an image stream between different locations using rostopic - local, urban, Germany-Spain, Germany-Ghana}
\label{rostopic_bw}
\end{figure*}


With regard to the ping test, we did not collect meaningful data for remote.it for a combination of reasons. remote.it relies on a cluster of proxy servers and the ping utility uses ICMP, a protocol which does not depend on ports of the target hosts. It is therefore not possible  to ping our intended targets through just one port, as one would end up pinging the proxy server cluster. Out of the remaining 3 methods, CJDNS had the highest minimum, maximum and average round trip times for Bremen Airport and Girona, but performed surprisingly well for the Ghana tests. Husarnet and Yggdrasil were in close competition for the fastest round trip times, with Husarnet having a slim lead. This could be due to differences in the routing protocol created by Yggdrasil as opposed to that of Husarnet, which resulted in a few packages being delayed a little longer before reaching their destination. The ping times did seem to be affected by distance, since the minimum round trip times were higher for the Girona tests than the downtown tests and were highest for the Ghana test, which was the farthest geographically.

Focusing on the iperf tests, UDP transmission speed was fairly constant around 1.0 Mbps except in the Ghana tests, while TCP transmission speeds went as high as 40 Mbps. In some cases, upload speed was higher than download speed. We attribute this to the network architecture of Jacobs University Bremen and the configuration of its ISP, the Deutsche Forschungsnetz. Since only some of the results were affected by this trend, it lends to the idea of a more non-deterministic effect on the incoming and outgoing transmissions, which we are currently not fully able to describe. But overall, the results show that the networks have high potential to support transmission of large data at a reasonably fast rate, which is commonplace when working with robotics systems.

Concerning SSH, we again noticed the non-deterministic trend that upload speeds were higher than download speeds. Husarnet was almost as fast as Yggdrasil when we tested within the same city (Bremen), but did not scale as well as Yggdrasil, which supported the long distance connections to Girona and Ghana better. Objectively, working over ssh over the 4 networking methods, there was not much noticeable lag between the typing of the commands and their being sent to the remote host, although this did happen once or twice on days with bad connections, the lag was not for more than 1-2 seconds. In the case of the intercontinental connection, the lag was much more noticeable. There were infrequent dropouts, the number of which grew on days when we tested with a more unstable Internet connection. In such cases a simple re-connection after some time fixed most problems. 

Obtaining the data from the \texttt{rostopic bw} command was challenging, since the raw output from the command itself did not strictly follow any format, and we could not statistically analyse it to generate graphs. There is also no option from the command to output in CSV, hence requiring an additional means to parse the data. We used a Python3 script to read and convert the data to CSV format. The image stream over ROS was better carried by Yggdrasil for the downtown and campus tests, but for the Girona and Ghana tests, Husarnet was better by a slim margin. Again, we noticed the influence of the higher upload speeds and lower download speeds in the bandwidth measurements.

\section{Conclusions and Future Work}
\label{s:cfw}
After the analysis of the different results, there was no clear winner overall, since some networks were better in some aspects and not as good in others. However, they all show potential as alternatives for Port Forwarding since they support end to end encryption and can transfer large amounts of data at reasonable speeds, even over distances of more than 6700 km.  
In general, the trend of upload speeds being higher than download speeds heavily influenced the results, since the campus network of Jacobs University was involved in all the tests. There is a possibility that the results could be different if a different internal network architecture was used at the Jacobs University end of the tests. 
There are several aspects which would still be worth exploring.
In the future, some areas to look at include: more extensive tests comparing different messaging frameworks such as ROS, ROS2, ZeroMQ; tests to quantitatively ascertain how feasible tele-operation is on these networks as well as field tests with robots connected via mobile carriers instead of Wifi.

\section*{Acknowledgements}
The authors would like to thank the Jacobs Robotics community, the organisers of the Jacobs Robotics Summer Academy where the initial part of this work was developed, and the Computer Vision and Robotics Institute (VICOROB) at University of Girona for their help and support.

\bibliographystyle{IEEEtran}
\bibliography{bib}

\begin{thebibliography}{10}
\providecommand{\url}[1]{#1}
\csname url@samestyle\endcsname
\providecommand{\newblock}{\relax}
\providecommand{\bibinfo}[2]{#2}
\providecommand{\BIBentrySTDinterwordspacing}{\spaceskip=0pt\relax}
\providecommand{\BIBentryALTinterwordstretchfactor}{4}
\providecommand{\BIBentryALTinterwordspacing}{\spaceskip=\fontdimen2\font plus
\BIBentryALTinterwordstretchfactor\fontdimen3\font minus
  \fontdimen4\font\relax}
\providecommand{\BIBforeignlanguage}[2]{{%
\expandafter\ifx\csname l@#1\endcsname\relax
\typeout{** WARNING: IEEEtran.bst: No hyphenation pattern has been}%
\typeout{** loaded for the language `#1'. Using the pattern for}%
\typeout{** the default language instead.}%
\else
\language=\csname l@#1\endcsname
\fi
#2}}
\providecommand{\BIBdecl}{\relax}
\BIBdecl

\bibitem{ifr2018}
I.~F. of~Robotics, ``Robot density rises globally,''
  \url{https://ifr.org/news/robot-density-rises-globally/}, 2018, [Online;
  accessed 2021-04-16].

\bibitem{Vernon2019}
D.~{Vernon}, ``Robotics and artificial intelligence in africa [regional],''
  \emph{IEEE Robotics Automation Magazine}, vol.~26, no.~4, pp. 131--135, 2019.

\bibitem{Gaus2019}
A.~Gaus and W.~Hoxtell, \emph{Automation and the Future of Work in Sub-Saharan
  Africa}.\hskip 1em plus 0.5em minus 0.4em\relax Sankt Augustin and Berlin,
  Germany: Konrad-Adenauer-Stiftung e.V., 2019.

\bibitem{Birk2021_edu}
A.~Birk, E.~Dineva, F.~Maurelli, and A.~Nabor, ``A robotics course during
  covid-19: Lessons learned and best practices for online teaching beyond the
  pandemic,'' \emph{Robotics}, vol.~10, no.~1, 2021.

\bibitem{Maurelli2021_rie}
F.~Maurelli, E.~Dineva, A.~Nabor, and A.~Birk, ``Robotics and intelligent
  systems: a new curriculum development and adaptations needed in coronavirus
  times,'' in \emph{Proceedings of the 12th International Conference on
  Robotics in Education}, 2021.

\bibitem{Fitz1999}
T.~Fitzpatrick, ``Live remote control of a robot via the internet,'' \emph{IEEE
  Robotics Automation Magazine}, vol.~6, no.~3, pp. 7--8, 1999.

\bibitem{saucy2000}
P.~Saucy and F.~Mondada, ``Khepontheweb: open access to a mobile robot on the
  internet,'' \emph{IEEE Robotics Automation Magazine}, vol.~7, no.~1, pp.
  41--47, 2000.

\bibitem{Luo2000}
R.~Luo and T.~M. Chen, ``Development of a multi-behavior based mobile robot for
  remote supervisory control through the internet,'' \emph{IEEE/ASME
  Transactions on Mechatronics}, vol.~5, no.~4, pp. 376--385, 2000.

\bibitem{Han2001}
K.-H. Han, S.~Kim, Y.-J. Kim, and J.-H. Kim, ``Internet control architecture
  for internet-based personal robot,'' \emph{Autonomous Robots}, vol.~10, p.
  135–147, 2001.

\bibitem{Wang2006}
Q.~Wang, S.~Liu, and Z.~Wang, ``A new internet architecture for robot remote
  control,'' in \emph{2006 IEEE/RSJ International Conference on Intelligent
  Robots and Systems}, 2006, pp. 4989--4993.

\bibitem{Nad2014}
J.~Nádvorník and P.~Smutný, ``Remote control robot using android mobile
  device,'' in \emph{Proceedings of the 2014 15th International Carpathian
  Control Conference (ICCC)}, 2014, pp. 373--378.

\bibitem{He2018}
H.~Su, J.~Li, K.~Kong, and J.~Li, ``Development and experiment of the
  internet-based telesurgery with microhand robot,'' \emph{Advances in
  Mechanical Engineering}, vol.~10, no.~2, p. 1687814018761921, 2018.

\bibitem{Tsokalo2019}
I.~A. Tsokalo, D.~Kuss, I.~Kharabet, F.~H.~P. Fitzek, and M.~Reisslein,
  ``Remote robot control with human-in-the-loop over long distances using
  digital twins,'' in \emph{2019 IEEE Global Communications Conference
  (GLOBECOM)}, 2019, pp. 1--6.

\bibitem{Miao2018}
Y.~Miao, Y.~Jiang, L.~Peng, M.~S. Hossain, and G.~Muhammad, ``Telesurgery robot
  based on 5g tactile internet,'' \emph{Mobile Networks and Applications},
  vol.~23, p. 1645–1654, 2018.

\bibitem{ROS-PW}
S.~S.~H. Hajjaj and K.~S.~M. Sahari, ``Establishing remote networks for ros
  applications via port forwarding: A detailed tutorial,'' \emph{International
  Journal of Advanced Robotic Systems}, vol.~14, no.~3, 2017.

\bibitem{cjdns}
C.~J. DeLisle, ``cjdns - networking reinvented,''
  \url{https://github.com/cjdelisle/cjdns}, 2020, [Online; accessed
  2021-04-16].

\bibitem{yggdrasil}
``Yggdrasil - end-to-end encrypted ipv6 networking to connect worlds,''
  \url{https://yggdrasil-network.github.io/}, 2020, [Online; accessed
  2021-04-16].

\bibitem{husarnet}
``husarnet - one network for all devices,'' \url{https://husarnet.com/},
  [Online; accessed 2021-04-16].

\bibitem{remote.it}
``remote.it - trusted access for private networks,'' \url{https://remote.it/},
  [Online; accessed 2021-04-16].

\end{thebibliography}

\end{document}